\documentclass[11pt,letterpaper]{article}

\usepackage{authblk}
\usepackage[hyperref]{naaclhlt2019}
\usepackage{pslatex}
\usepackage[english]{babel}
\usepackage[utf8]{inputenc}
\usepackage{amsmath}
\usepackage{bm}
\usepackage{graphicx}
\usepackage{tikz}
\usepackage{xcolor}
\usepackage{url}
\usepackage{rotating}
\usepackage{natbib}
\usepackage{amssymb}
\usepackage{linguex}
\usepackage{xspace}
\usepackage{subfig}
\usepackage{placeins}
\usepackage{pifont}
\usepackage{soul}

\newcommand{\cmark}{\ding{51}}%
\newcommand{\xmark}{\ding{55}}%

\newcommand{\key}[1]{\textbf{#1}}
\newcommand{\soft}[1]{}
\newcommand{\nopreview}[1]{}

\newcommand{\gulordavarnn}{GRNN\xspace}
\newcommand{\googlernn}{JRNN\xspace}

\newcommand{\tinylstm}{TinyLSTM\xspace}
\newcommand{\rnng}{RNNG\xspace}
\usepackage{xcolor}

\aclfinalcopy

\title{Neural Language Models as Psycholinguistic Subjects: Representations of Syntactic State}
\author[1]{\textbf{Richard Futrell}}
\author[2]{\textbf{Ethan Wilcox}}
\author[3,4]{\textbf{Takashi Morita}}
\author[5]{\textbf{Peng Qian}}
\author[6]{\textbf{Miguel Ballesteros}}
\author[5]{\textbf{Roger Levy}}

\affil[1]{Department of Language Science, UC Irvine, \tt{rfutrell@uci.edu}}
\affil[2]{Department of Linguistics, Harvard University, \tt{wilcoxeg@g.harvard.edu}}
\affil[3]{Primate Research Institute, Kyoto University, \tt{tmorita@alum.mit.edu}}
\affil[4]{Department of Linguistics and Philosophy, MIT}
\affil[5]{Department of Brain and Cognitive Sciences, MIT, \tt{rplevy@mit.edu}}
\affil[6]{IBM Research, MIT-IBM Watson AI Lab, \tt{miguel.ballesteros@ibm.com}}

\date{}

\begin{document}

\maketitle

\begin{abstract} 
We deploy the methods of controlled psycholinguistic experimentation to shed light on the extent to which the behavior of neural network language models reflects incremental representations of syntactic state. To do so, we examine model behavior on artificial sentences containing a variety of syntactically complex structures. We test four models: two publicly available LSTM sequence models of English \citep{jozefowicz2016exploring,gulordava2018colorless} trained on large datasets; an RNNG \citep{dyer2016recurrent} trained on a small, parsed dataset; and an LSTM trained on the same small corpus as the RNNG. We find evidence that the LSTMs trained on large datasets represent syntactic state over large spans of text in a way that is comparable to the RNNG, while the LSTM trained on the small dataset does not or does so only weakly.
\end{abstract}


\section{Introduction}
It is now standard practice in NLP to derive sentence representations using neural sequence models of various kinds \citep{elman1990finding,sutskever2014sequence,goldberg2017neural,peters2018deep,devlin2018bert}. However, we do not yet have a firm understanding of the precise content of these representations, which poses problems for interpretability, accountability, and controllability of NLP systems. More specifically, the success of neural sequence models has raised the question of whether and how these networks learn robust syntactic generalizations about natural language, which would enable robust performance even on data that differs from the peculiarities of the training set.

Here we build upon recent work studying neural language models using experimental techniques that were originally developed in the field of psycholinguistics to study language processing in the human mind. The basic idea is to examine language models' behavior on targeted sentences chosen to probe particular aspects of the learned representations. This approach was introduced by \citet{linzen2016assessing}, followed more recently by others \citep{bernardy2017using,enguehard2017exploring,gulordava2018colorless}, who used an agreement prediction task \citep{bock1991broken} to study whether RNNs learn a hierarchical morphosyntactic dependency: for example, that \emph{The key to the cabinets\dots} can grammatically continue with \emph{was} but not with \emph{were}.  This dependency turns out to be learnable from a language modeling objective \citep{gulordava2018colorless}. Subsequent work has extended this approach to other grammatical phenomena, with positive results for filler--gap dependencies \citep{chowdhury2018rnn,wilcox2018what} and negative results for anaphoric dependencies \citep{marvin2018targeted}.

In this work, we consider syntactic representations of a different kind. Previous studies have focused on relationships of dependency: one word licenses another word, which is tested by asking whether a language model favors one (grammatically licensed) form over another in a particular context. 
Here we focus instead on whether neural language models show evidence for incremental \key{syntactic state} representations: whether behavior of neural language models reflects the kind of generalizations that a symbolic grammar-based description of language would capture using a stack-based incremental parse state. For example, during the underlined portion of Example~\ref{ex:subordination-basic}, an incremental language model should represent and maintain the knowledge that it is currently inside a subordinate clause, implying (among other things) that a full main clause must follow.

\ex. \label{ex:subordination-basic} As \ul{the doctor studied the textbook}, the nurse walked into the office.

In this work, we use a targeted evaluation approach \citep{marvin2018targeted} to elicit evidence for syntactic state representations from language models. That is, we examine language model behavior on artificially constructed sentences designed to expose behavior that is crucially dependent on syntactic state representations. In particular, we study complex subordinate clauses and garden path effects (based on main-verb/reduced-relative ambiguities and NP/Z ambiguities). We ask three general questions: (1) Is there basic evidence for the representation of syntactic state? (2) What textual cues does a neural language model use to infer the beginnings and endings of such states? (3) Do the networks maintain knowledge about syntactic states over long spans of complex text, or do the syntactic state representations degrade?

\begin{table*}
    \centering
    \begin{tabular}{|l|l|l|r|l|}
        \hline
        Model & Architecture & Training data & Data size (tokens) & Reference\\
        \hline
        \googlernn & LSTM & One Billion Word & $\sim 800$ million & \citet{jozefowicz2016exploring} \\
         \gulordavarnn & LSTM & Wikipedia & $\sim 90$ million & \citet{gulordava2018colorless} \\
         \rnng & RNN Grammar & Penn Treebank & $\sim 1$ million & \citet{dyer2016recurrent} \\
         \tinylstm & LSTM & Penn Treebank & $\sim 1$ million & --- \\
         \hline
    \end{tabular}
    \caption{Models tested, by architecture, training data, and training data size.}
    \label{tab:models}
\end{table*}

Among neural language models, we study both generic sequence models (LSTMs), which have no explicit representation of syntactic structure, and an RNN Grammar (RNNG) \citep{dyer2016recurrent}, which explicitly calculates Penn Treebank-style context-free syntactic representations as part of the process of assigning probabilities to words. This comparison allows us to evaluate the extent to which explicit representation of syntactic structure makes models more or less sensitive to syntactic state. RNNGs have been found to outperform LSTMs not only in overall test-set perplexity \citep{dyer2016recurrent}, but also in modeling long-distance number agreement in \citet{kuncoro2018lstms} for certain model configurations; our work extends this comparison to a variety of syntactic state phenomena. 

\section{General methods}

We investigate neural language model behavior primarily by studying the \key{surprisal}, or log inverse probability, that a language model assigns to each word in a sentence:
\begin{equation*}
S(x_i) = -\log_2 p(x_i|h_{i-1}),
\end{equation*}
where $x_i$ is the current word or character, $h_{i-1}$ is the model's hidden state before consuming $x_i$, the probability is calculated from the network's softmax activation, and the logarithm is taken in base 2, so that surprisal is measured in bits. Surprisal is equivalent to the pointwise contribution to the language modeling loss function due to a word.

In psycholinguistics, the common practice is to study reaction times per word (for example, reading time as measured by an eyetracker), as a measure of the word-by-word difficulty of online language processing. These reading times are often taken to reflect the extent to which humans expect certain words in context, and may be generally proportional to surprisal given the comprehender's probabilistic language model \citep{hale2001probabilistic,levy2008expectation,smith2013effect}. In this study, we take language model surprisal as the analogue of human reading time, using it to probe the neural networks' expectations about what words will follow in certain contexts. There is a long tradition linking RNN performance to human language processing \citep{elman1990finding,christiansen1999toward,macdonald-christiansen:2002} and grammaticality judgments \citep{lau2017grammaticality}, and RNN surprisals are a strong predictor of human reading times \citep{frank2011insensitivity,goodkind2018predictive}. RNNGs have also been used as models of human online language processing \citep{hale2018finding}.

\subsection{Experimental methodology}

In each experiment presented below, we design a set of sentences such that the word-by-word surprisal values will show evidence for syntactic state representations. The idea is that certain words will be surprising to a language model only if the model has a representation of a certain syntactic state going into the word. We analyze word-by-word surprisal profiles for these sentences using regression analysis. Except where otherwise noted, all statistics are derived from linear mixed-effects models \citep{baayen2008mixed} with sum-coded fixed-effect predictors and maximal random slope structure \citep{barr2013random}. This method lets us factor out by-item variation in surprisal and focus on the contrasts between conditions.

\subsection{Models tested}

We study the behavior of four models of English: two LSTMs trained on large data, an an RNNG and an LSTM trained on matched, smaller data (the Penn Treebank). The models are summarized in Table~\ref{tab:models}. All models are trained on a language modeling objective. 

Our first LTSM is the model presented in \citet{jozefowicz2016exploring} as ``BIG LSTM+CNN Inputs'', which we call ``\googlernn'', which was trained on the One Billion Word Benchmark \citep{chelba2013one} with two hidden layers of 8196 units each and CNN character embeddings as input. The second large LSTM is the model described in the supplementary materials of \citet{gulordava2018colorless}, which we call ``\gulordavarnn'', trained on 90 million tokens of English Wikipedia with two hidden layers of 650 hidden units each. 

Our \rnng is trained on syntactically labeled Penn Treebank data \citep{marcus1993building}, using 256-dimensional word embeddings for the input layer and 256-dimensional hidden layers, and dropout probability 0.3. Next-word predictions are obtained through hierarchical softmax (140 clusters, obtained with the greedy agglomerative clustering algorithm of \newcite{Brown:1992}). We estimate word surprisals using word-synchronous beam search  \cite{stern-etal:2017effective-inference,hale2018finding}: at each word $w_i$ a beam of incremental parses is filled, the summed forward probabilities \citep{stolcke:1995} of all candidates on the beam is taken as a lower bound on the prefix probability: $P_{\text{min}}(w_{1\dots i})$, and the surprisal of the $i$-th word in the sentence is estimated as $\log \frac{P_{\text{min}}(w_{1\dots i})}{P_{\text{min}}(w_{1\dots i-1})}$.  Our action beam is size 100, and our word beam is size 10.
Finally, we use an LSTM trained on string data from the Penn Treebank training set, which we call \tinylstm, to disentangle effects of training set from model architecture.  For \tinylstm we use 256-dimensional word-embedding inputs and hidden layers and dropout probability 0.3, just as with the \rnng. 

\section{Subordinate clauses}
\label{sec:obligatory-upcoming}

\begin{figure}[!ht]
\centering
\includegraphics[scale=.7]{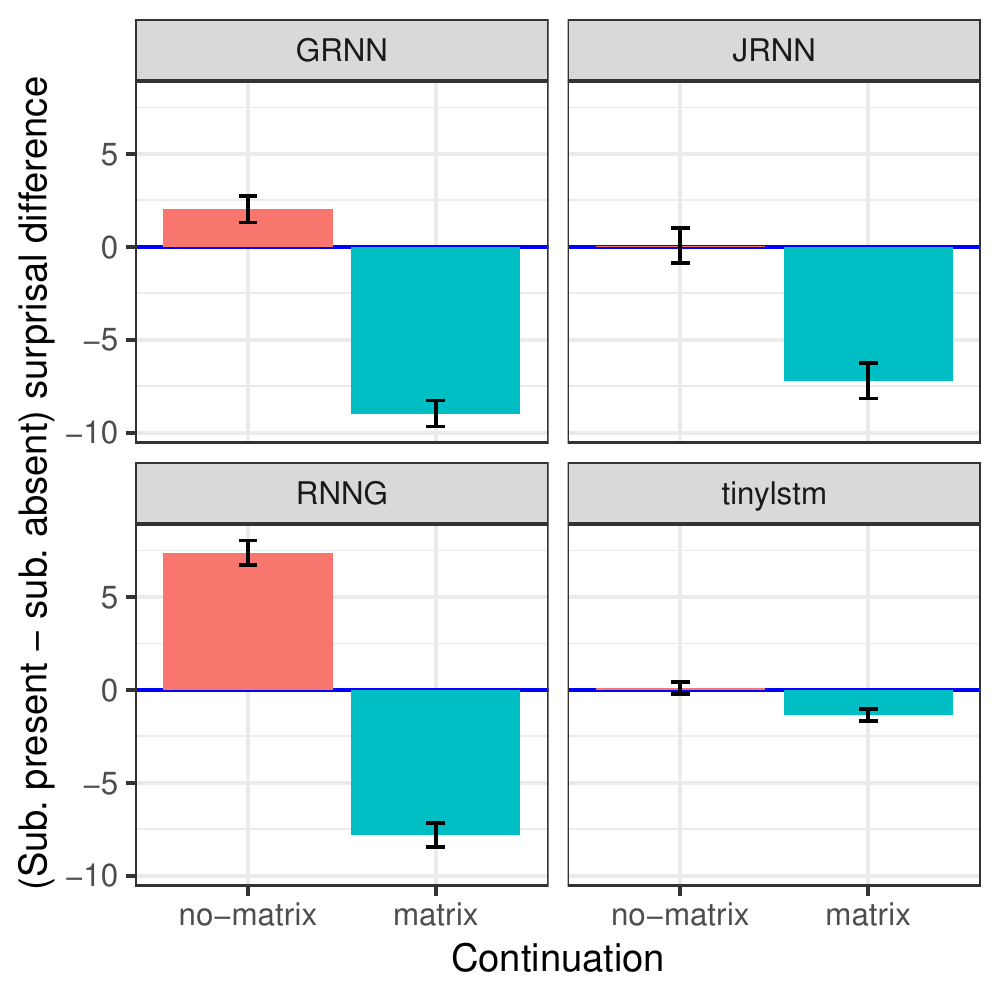}
\caption{Effect of subordinator absence/presence on surprisal of continuations. Red: no-matrix penalty effect. Blue: matrix licensing effect. In this and all other figures, unless otherwise noted, \key{error bars} represent 95\% confidence intervals of the \emph{contrasts between conditions shown}, computed from the standard error of the by-item and by-condition mean surprisals after subtracting out the by-item means \citep{masson2003using}.}
\label{fig:basic-subordination}
\end{figure}

We begin by studying subordinate clauses, a key example of a construction requiring stack-like representation of syntactic state. In such constructions, as shown in Example~\ref{ex:subordination-basic}, a \key{subordinator} such as ``as'' or ``when'' serves as a cue that the following clause is a subordinate clause, meaning that it must be followed by some main (matrix) clause. In an incremental language model, this knowledge must be maintained and carried forward while processing the words inside subordinate clause. A grammar-based symbolic language model \citep[e.g.,][]{stolcke:1995,manning2000probabilistic} would maintain this knowledge by keeping track of syntactic rules representing the incomplete subordinate clause and the upcoming main clause in a stack data structure. Psycholinguistic research has clearly demonstrated that humans maintain representations of this kind in syntactic processing \citep{staub-clifton:2006,lau-etal:2006,levy-fedorenko-breen-gibson:2012cognition}. Here we ask whether the string completion probabilities produced by neural language models show evidence of the same knowledge.

We can detect the knowledge of syntactic state in this case by examining whether the network licenses and requires a matrix clause following the subordinate clause. These expectations can be detected by examining surprisal differences between sentences of the form in Example~\ref{ex:subordination}:
{\setlength{\Exlabelwidth}{1em}
\setlength{\Exlabelsep}{1em}
\ex. \label{ex:subordination}
\a. As the doctor studied the textbook, the nurse walked into the office. [\textsc{sub}ordinator, \textsc{matrix}]\label{ex:subordination-sub-matrix}
\b. *As the doctor studied the textbook. [\textsc{sub}, \textsc{no-matrix}] \label{ex:subordination-sub-nomatrix}
\c. ?The doctor studied the textbook, the nurse walked into the office. [\mbox{\textsc{no-sub}ordinator},~\textsc{matrix}] \label{ex:subordination-nosub-matrix}
\d. The doctor studied the textbook. [\mbox{\textsc{no-sub}}, \textsc{no-matrix}] \label{ex:subordination-nosub-nomatrix}
\z.

}

\begin{figure*}[!ht]
\centering
\begin{minipage}{0.4\textwidth}
\includegraphics[width=\textwidth]{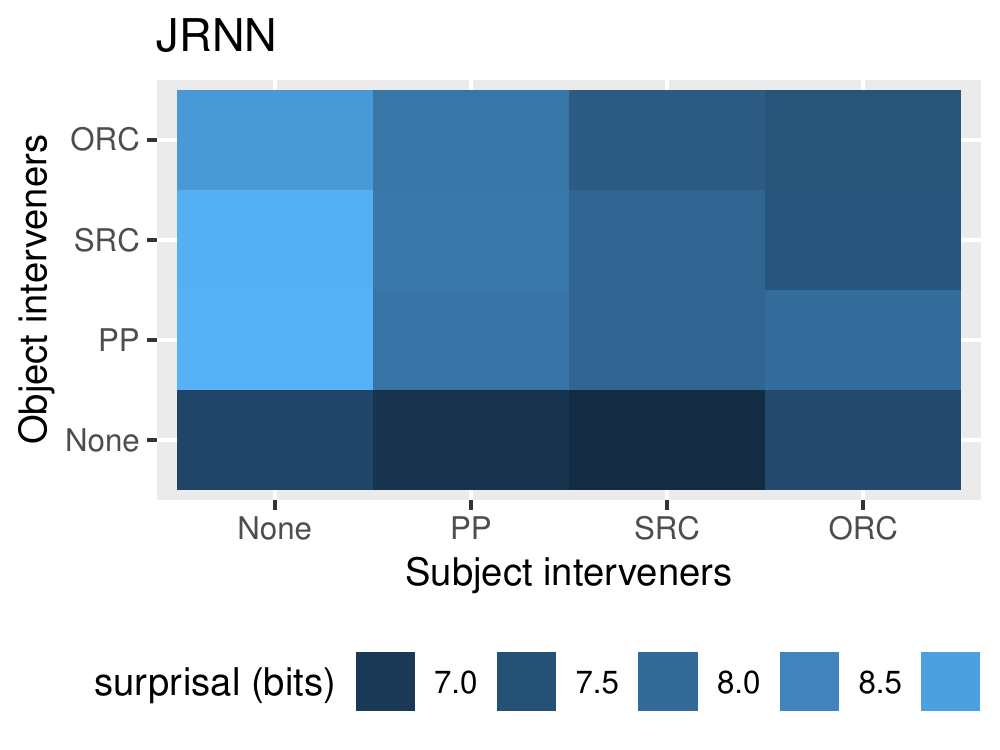}
\end{minipage}
\begin{minipage}{0.4\textwidth}
\includegraphics[width=\textwidth]{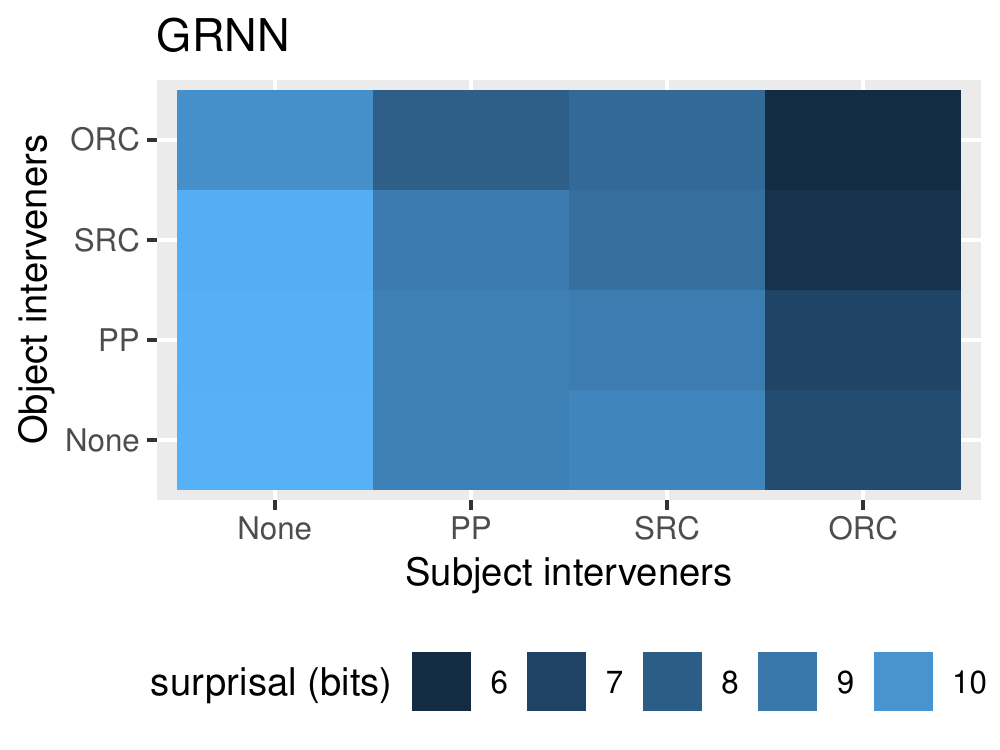}
\end{minipage}\\
\begin{minipage}{0.4\textwidth}
\includegraphics[width=\textwidth]{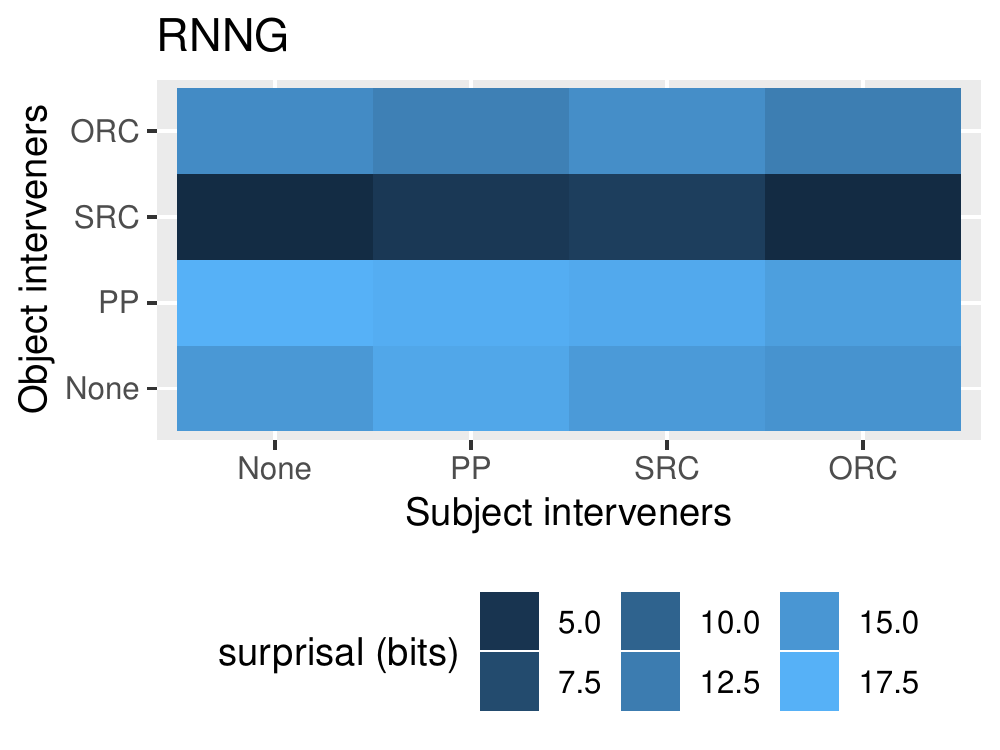}
\end{minipage}
\begin{minipage}{0.4\textwidth}
\includegraphics[width=\textwidth]{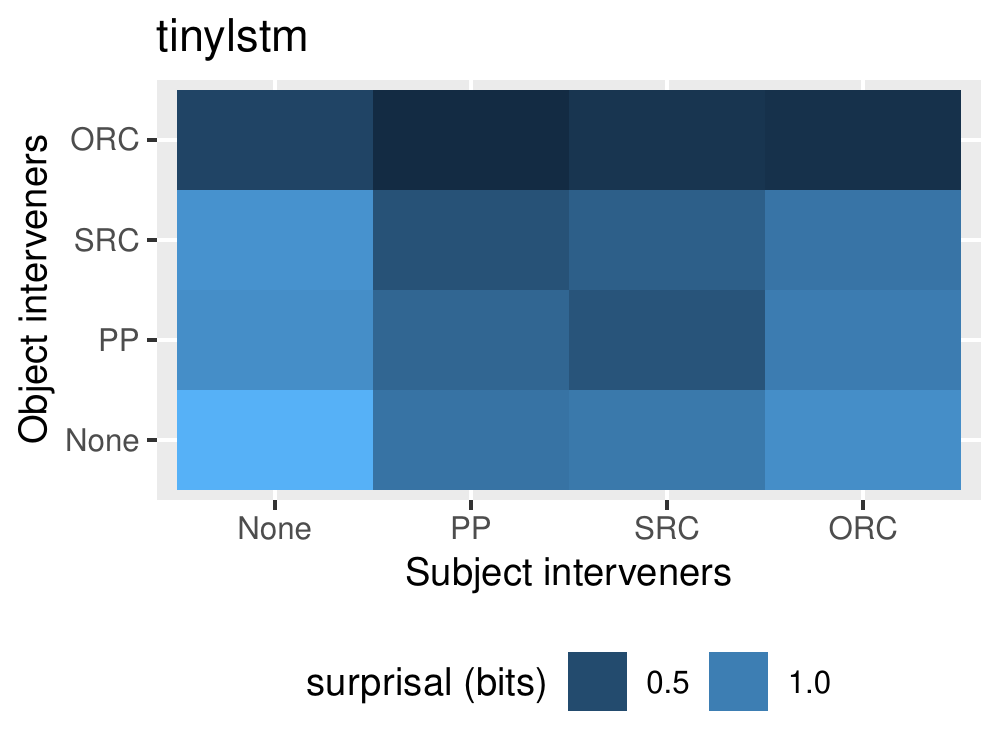} 
\end{minipage}

\vspace{-0.15cm}
    \caption{Size of matrix clause licensing interaction (see text) given various intervening elements in the subordinate clause. Note that the heatmaps are on different scales across models.}
\label{fig:subordination-results}
\end{figure*} 

If the network \emph{licenses} a matrix clause following the subordinate clause---and maintained knowledge of that licensing relationship throughout the clause, from the subordinator to the comma---then this should be manifested as lower surprisal at the matrix clause in \ref{ex:subordination-sub-matrix} as compared to \ref{ex:subordination-nosub-matrix}. We call this the \key{matrix licensing effect}: the surprisal of the condition [\textsc{sub}, \textsc{matrix}] minus [\textsc{nosub}, \textsc{matrix}], which will be negative if there is a licensing effect. If the network \emph{requires} a following matrix clause, then this will be manifested as higher surprisal at the matrix clause for \ref{ex:subordination-sub-nomatrix} compared with \ref{ex:subordination-nosub-nomatrix}. We call this the \key{no-matrix penalty effect}: the surprisal of [\textsc{sub},\textsc{nomatrix}] minus [\textsc{nosub}, \textsc{nomatrix}], which will be positive if there is a penalty.

We designed 23 experimental items on the pattern of \ref{ex:subordination} and calculated difference in the sum surprisal of the words in the matrix clause. Figure~\ref{fig:subordination-results} shows the matrix licensing effect (in blue) and the no-matrix penalty effect (in red), averaged across items. For all models, we see a facilitative matrix licensing effect ($p<.001$ for all models), smallest in \tinylstm. However, we only find a significant no-matrix penalty for \gulordavarnn and the \rnng ($p<.001$ in both): the other models do not significantly penalize an ungrammatical continuation ($p=.9$ for \googlernn; $p=.5$ for \tinylstm). That is, \googlernn and \tinylstm give no indication that \ref{ex:subordination-sub-nomatrix} is less probable than \ref{ex:subordination-nosub-matrix}.

We found that all models at least partially represent the licensing relationship between a subordinate and matrix clause. However, in order to fully represent the syntactic requirements induced by a subordinator, it seems that a model needs either large amounts of data (as in \gulordavarnn) or explicit representation of syntax (as in the \rnng, as opposed to \tinylstm). 

\subsection{Maintenance and degradation of syntactic state}
\label{sec:subordination-degradation}

The foregoing results show that neural language models use the presence of a subordinator as a cue to the onset of a subordinate clause, and that they maintain knowledge that they are in a subordinate clause throughout the intervening material up to the comma. Now we probe the ability of models to maintain this knowledge over long spans of complex intervening material. To do so, we use sentences on the template of \ref{ex:subordination} and add intervening material modifying the NPs in the subordinate clause. To both of these NPs (in subject and object position), we add modifiers of increasing syntactic complexity: PPs, subject-extracted relative clauses (SRCs), and object-extracted relative clauses (ORCs). We study the extent to which these modifiers weaken the language models' expectations about the upcoming matrix clause. 

As a summary measure of the strength of language models' expectations about an upcoming matrix clause, we collapse the two measures of the previous section into one: the \key{matrix licensing interaction}, consisting of the difference between the no-matrix penalty effect and the matrix licensing effect (the two bars in Figure~\ref{fig:basic-subordination}). A similar measure was used to detect filler--gap dependencies in \cite{wilcox2018what}. 

Figure~\ref{fig:subordination-results} shows the strength of the matrix licensing interaction given sentences with various modifiers inserted. For the large LSTMs, \gulordavarnn exhibits a strong interaction when the intervening material is short and syntactically simple, and the interaction gets progressively weaker as the intervening material becomes progressively longer and more complex ($p<0.001$ for subject postmodifiers and $p<0.01$ object postmodifiers). The other models show less interpretable behavior. 

Our results indicate that at least some large LSTMs, along with the \rnng, are capable of maintaining a representation of syntactic state over spans of complex intervening material. Quantified as a licensing interaction, this representation of syntactic state exhibits the most clearly understandable behavior in \gulordavarnn, which shows a graceful degradation of syntactic expectations as the complexity of intervening material increases. The representation is maintained most strongly in the \rnng, except for one particular construction (object-position SRCs).

\section{Garden path effects}
\label{sec:garden-path}

The major phenomenon that has been used to probe incremental syntactic representations in humans is \key{garden path effects}. Garden path effects arise from local ambiguities, where a context leads a comprehender to believe one parse is likely, but then a disambiguating word forces her to drastically revise her beliefs, resulting in high surprisal/reading time at the disambiguating word. In effect, the comprehender is ``led down the garden path'' by a locally likely but ultimately incorrect parse \citep{bever1970cognitive}. Garden-pathing in LSTMs has recently been demonstrated by \citet{vanschijndel2018modeling,vanschijndel2018neural} in the context of modeling human reading times.

Garden path effects allow us to detect representations of syntactic state because if a person or language model shows a garden path effect at a word, that means that the person or model had some belief about syntactic state which was disconfirmed by that word. In psycholinguistics, these effects have been used to study the question of what information determines people's beliefs about likely parses given locally ambiguous contexts: for example, whether factors such as world knowledge play a role \citep{ferreira1986independence,trueswell1994semantic}. 

Here we study two major kinds of local ambiguities inducing garden path effects. For each ambiguity, we ask two main questions. First, whether the network shows the basic garden path effect, which would indicate that it had a syntactic state representation that made a disambiguating word surprising. Second, whether the network is sensitive to subtle lexical cues to syntactic structure which may modulate the size of the garden path effect: this question allows us to determine what information the network uses to determine the beginnings and endings of certain syntactic states. 

\subsection{NP/Z Ambiguity}
\label{sec:npz}

The \key{NP/Z ambiguity}\footnote{For Noun Phrase/Zero ambiguity. At first the embedded verb appears to take an NP object, but later it turns out that it was a zero (null) object.} refers to a local ambiguity in sentences of the form given in Example~\ref{ex:npz-transitivity}. When a comprehender reads the underlined phrase ``\ul{the vet with his new assistant}'' in ~\ref{ex:npz-transitive}, she may at first believe that this phrase is the direct object of the verb ``scratched'' inside the subordinate clause. However, upon reaching the verb ``took off'', she realizes that the underlined phrase was not in fact an object of the verb ``scratched'', rather it was the subject of a new clause, and the subordinate clause ended after the verb ``scratched''. The key region of the sentence where the garden path disambiguation happens---called the \key{disambiguator}---is the phrase ``took off'', marked in bold.

{
\setlength{\Exlabelwidth}{0.5em}
\setlength{\Exlabelsep}{0.7em}
\setlength{\SubExleftmargin}{1em}
\ex. \label{ex:npz-transitivity}
\a. When the dog scratched \ul{the vet with his new assistant} \textbf{took off} the muzzle. [\textsc{transitive}, \textsc{nocomma}]\label{ex:npz-transitive}
\b. When the dog scratched, \ul{the vet with his new assistant} \textbf{took off} the muzzle. [\textsc{transitive}, \textsc{comma}]\label{ex:npz-transitive-comma}
\c. When the dog struggled \ul{the vet with his new assistant} \textbf{took off} the muzzle. [\textsc{intransitive}, \textsc{nocomma}]\label{ex:npz-intransitive}
\d. When the dog struggled, \ul{the vet with his new assistant} \textbf{took off} the muzzle. [\textsc{intransitive}, \textsc{comma}]\label{ex:npz-intransitive-comma}  
\z.

}

While a garden path should obtain in ~\ref{ex:npz-transitive}, no such garden path should exist for ~\ref{ex:npz-transitive-comma}, because a comma clearly demarcates the end of the subordinate clause. Therefore a basic garden path effect would be indicated by the difference in surprisal at the disambiguator for ~\ref{ex:npz-transitive} minus ~\ref{ex:npz-transitive-comma}. Furthermore, if a comprehender is sensitive to the relationship between verb argument structure and clause boundaries, then there should be no garden path in ~\ref{ex:npz-intransitive}, because the verb ``struggled'' is \textsc{intransitive}: it cannot take an object in English, so an incremental parser should never be misled into believing that ``\ul{the vet...}'' is its object. This lexical information about syntactic structure is subtle enough that there has been controversy about whether even humans are sensitive to it in online processing \citep{staub2007parser}.

\subsubsection{NP/Z Garden Path Effect}

\begin{figure}
    \centering
    \includegraphics[scale=.7]{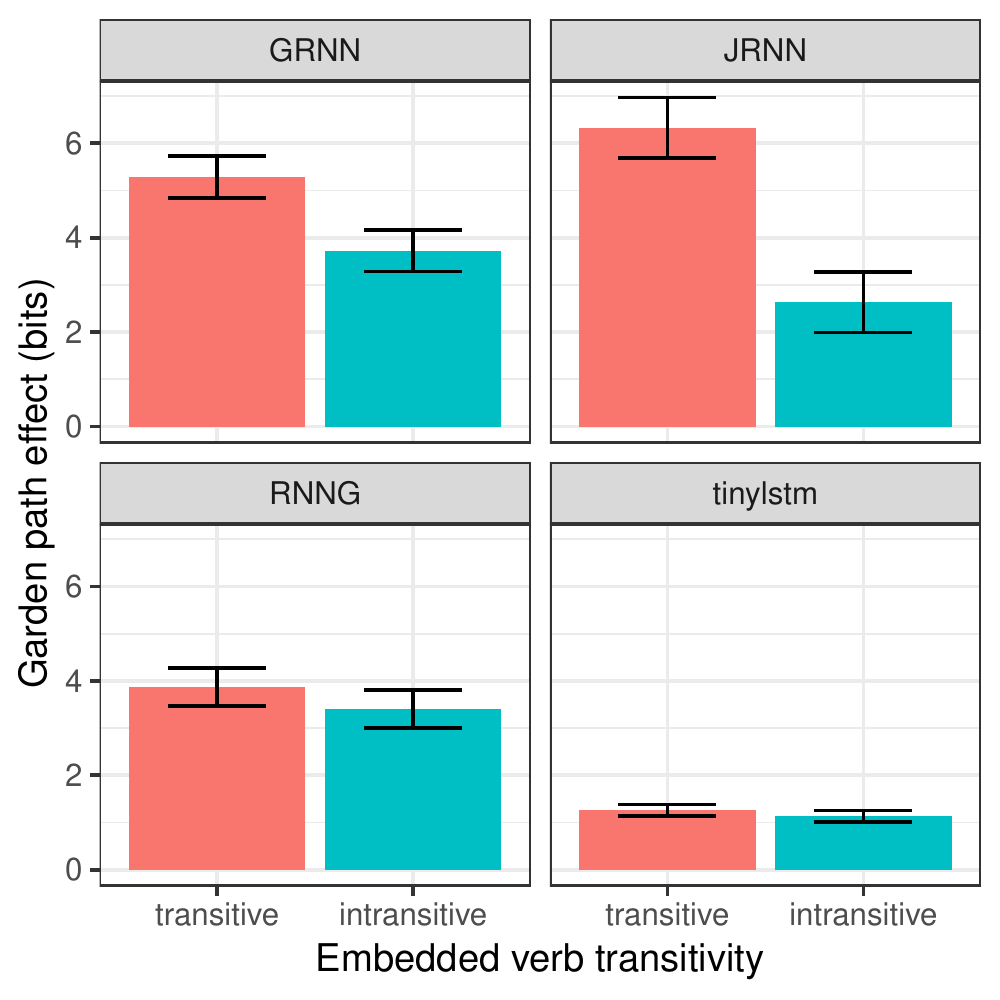}
    \caption{Average garden path effect (surprisal at disambiguator in \textsc{no-comma} condition minus \textsc{comma} condition) by model and embedded verb transitivity.}
    \label{fig:npz-transitivity}
\end{figure}

We tested whether neural language models would show the basic garden path effect and if this effect would be modulated by verb transitivity. We constructed 32 items based of the same structure as \ref{ex:npz-transitivity}, based on materials from \citet{staub2007parser}, manipulating the transitivity of the embedded verb (``scratched'' vs. ``struggled''), and the presence of a disambiguating comma at the end of the subordinate clause. An NP/Z garden path effect would show up as increased surprisal at the main verb ``took off'' in the absence of a comma. If the networks use the transitivity of the embedded verb as a cue to clause structure, and maintain that information over the span of six words between the embedded verb and the main verb, then there should be a garden path effect for the transitive verb, but not for the intransitive verb. More generally we would expect a \emph{stronger} garden path given the transitive verb than given the intransitive verb.

Figure~\ref{fig:npz-transitivity} shows the mean surprisals at the disambiguator for all four models, for both transitive and intransitive embedded verbs. We see that a garden path effect exists in all models (though very small in \tinylstm): all models show significantly higher surprisal at the main verb when the disambiguating comma is absent ($p<.001$ for all models). However, only the large LSTMs appear to be sensitive to the transitivity of the embedded verb, showing a smaller garden path effect for intransitive verbs. Statistically, there is a significant interaction of comma presence and verb transitivity only in \gulordavarnn and \googlernn (\gulordavarnn: $p<.01$; \googlernn: $p<.001$; \rnng: $p=.3$, \tinylstm: $p=.3$).

All models show NP/Z garden path effects, indicating that they are sensitive to some cues indicating end-of-clause boundaries. However, only the large LSTMs appear to use verb argument structure information as a cue to these boundaries. The results suggest that very large amounts of data may be necessary for current neural models to discover such fine-grained dependencies between syntactic properties of verbs and sentence structure.

\subsubsection{Maintenance and degradation of state}

\begin{figure}
    \centering
    \includegraphics[scale=.7]{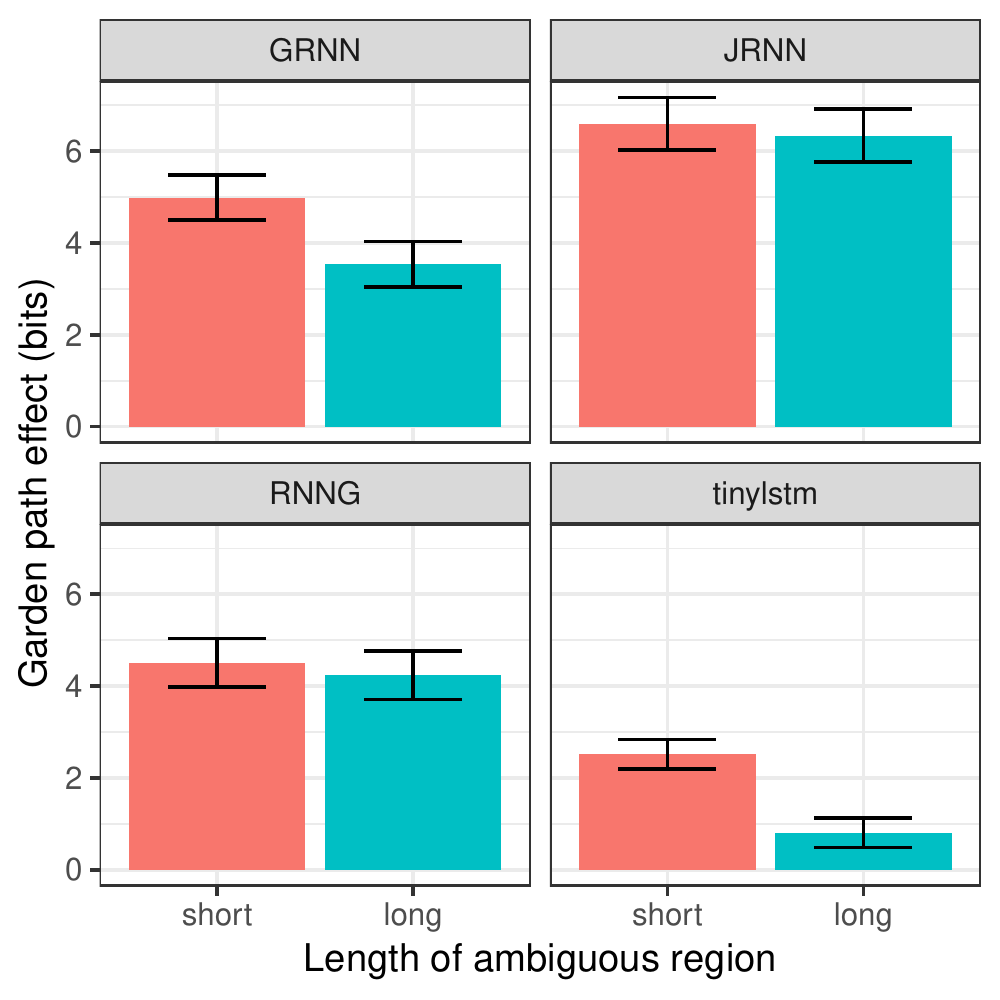}
    \caption{Average garden path effect by model and length of ambiguous region.}
    \label{fig:digging-in}
\end{figure}

We can probe the maintenance and degradation of syntactic state information by manipulating the length of the intervening material between the onset of the local ambiguity and the disambiguator in examples such as \ref{ex:npz-transitivity}. The question is whether the networks maintain the knowledge, while processing the intervening material, that the intervening noun phrase is probably the object of the embedded verb inside a subordinate clause, or whether they gradually lose track of this information. To study this question we used materials on the pattern of ~\ref{ex:digging-in}: these materials manipulate the length of the intervening material (underlined) while holding constant the distance between the subordinator (``As'') and the disambiguator (\textbf{grew}).

{
\setlength{\Exlabelwidth}{0.5em}
\setlength{\Exlabelsep}{0.7em}
\setlength{\SubExleftmargin}{1em}
\ex. \label{ex:digging-in}
\a. As the author studying Babylon in ancient times wrote \ul{the book} \textbf{grew}. [\textsc{short}, \textsc{nocomma}]\label{ex:di-short}
\b. As the author studying Babylon in ancient times wrote, \ul{the book} \textbf{grew}. [\textsc{short}, \textsc{comma}]\label{ex:di-short-comma}
\c. As the author wrote \ul{the book describing Babylon in ancient times} \textbf{grew}. [\textsc{long}, \textsc{nocomma}]\label{ex:di-long}
\d. As the author wrote, \ul{the book describing Babylon in ancient times} \textbf{grew}. [\textsc{long}, \textsc{comma}]\label{ex:di-long-comma}
\z.

} 

If neural language models show degradation of syntactic state, then the garden path effect (measured as the difference in surprisal between the \textsc{comma} and \textsc{no-comma} conditions at the disambiguator) will be smaller for the \textsc{long} conditions. We tested 32 sentences of the form in~\ref{ex:digging-in}, based on materials from \citet{tabor2004evidence}. The garden path effect sizes are shown in Figure~\ref{fig:digging-in}.

\begin{figure}
    \centering
    \includegraphics[scale=.7]{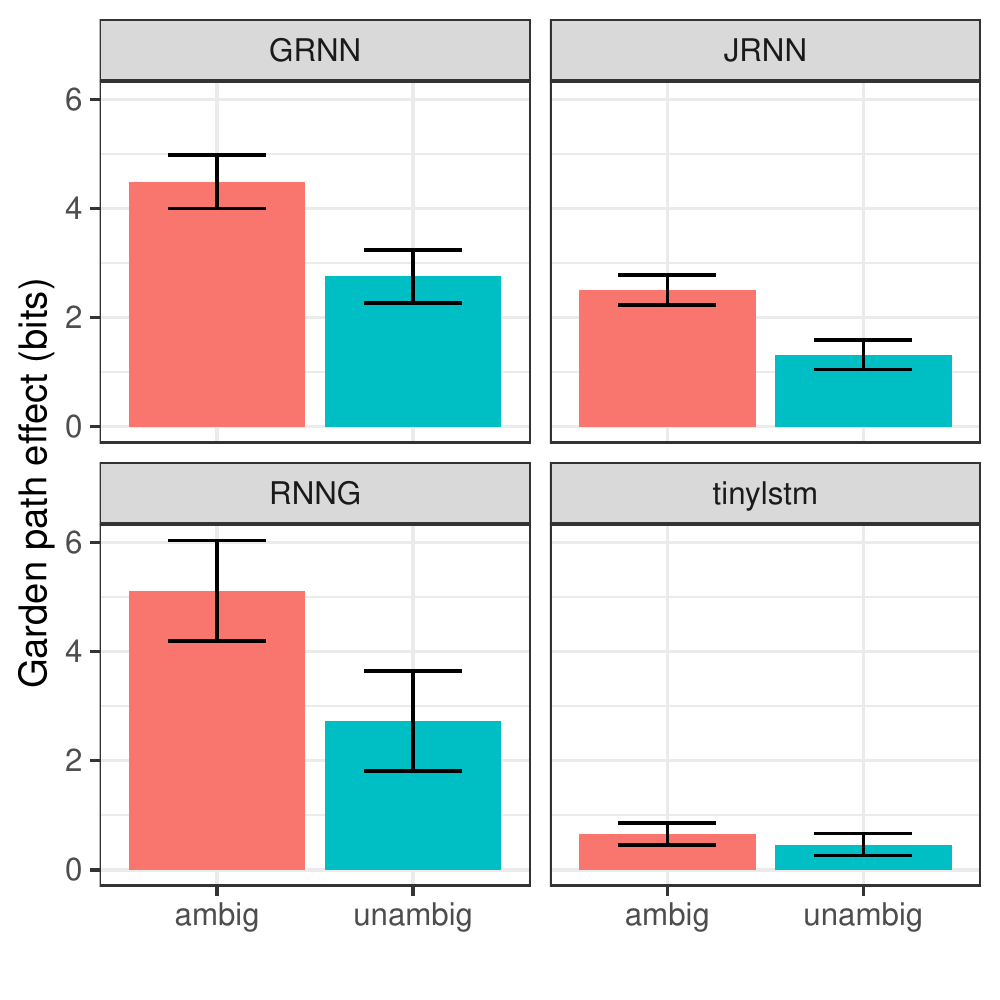}
    \caption{Garden path effect size for MV/RR ambiguity by model and verb-form ambiguity.}
    \label{fig:mv-rr-surprisal}
\end{figure}

We find a significant garden effect in all models in the \textsc{short} condition ($p<.001$ in \googlernn and \gulordavarnn; $p<.01$ in the RNNG and $p=.03$ in \tinylstm). In the long condition, we find the garden path effect in all models except \tinylstm: ($p<.001$ in \googlernn; $p<.01$ in \gulordavarnn; $p=.02$ in the \rnng; and $p=.2$ in \tinylstm). The crucial interaction between length and comma presence (indicating that syntactic state degrades) is significant in \gulordavarnn ($p<.01$) and \tinylstm ($p<.001$) but not \googlernn ($p=.7$) nor the \rnng ($p=.6$). The pattern is reminiscent of the results on degradation of state information about subordinate clauses in Section~\ref{sec:obligatory-upcoming}, where \gulordavarnn and \tinylstm showed the clearest evidence of degradation.

Note that the pattern found here is the opposite of the pattern of human reading times. Humans appear to show ``digging-in'' effects: the longer the span of time between the introduction of a local ambiguity and its resolution, the larger the garden path effect \citep{tabor2004evidence,levy2009modeling}. 

\begin{table*}[]
    \centering
    \begin{tabular}{|l|c|c|c|c|}
        \hline
         Phenomenon & \gulordavarnn & \googlernn & \rnng & \tinylstm \\
         \hline
         Subordination & \cmark \cmark & \cmark \xmark & \cmark \cmark & \cmark \xmark \\
         NP/Z Garden Path & \cmark \xmark & \cmark \cmark & \cmark \xmark & \cmark \xmark \\
         MV/RR Garden Path & \cmark \cmark & \cmark \xmark & \cmark \xmark & \cmark \xmark \\
         \hline
    \end{tabular}
    \caption{Summary of results by model and phenomenon. The first check mark indicates basic evidence of syntactic state representation. The second check mark indicates the ability to capture more fine-grained phenomena: for subordination, the no-matrix penalty effect; for the NP/Z garden path, the effect of verb transitivity; and for the MV/RR garden path, the effect of verb morphology.}
    \label{tab:overall-results}
\end{table*}

\subsection{Main Verb/Reduced Relative Ambiguity}
\label{sec:mv-rr}



Next we turn to garden path effects induced by the classic \key{Main Verb/Reduced Relative (MV/RR) ambiguity}, in which a word is locally ambiguous between being the main verb of a sentence or introducing a \key{reduced relative clause} (reduced RC: a relative clause with no explicit complementizer, headed by a passive-participle verb). That ambiguity can be maintained over a long stretch of material:

{
\setlength{\Exlabelwidth}{0.5em}
\setlength{\Exlabelsep}{0.7em}
\setlength{\SubExleftmargin}{1em}
\ex. \label{ex:mv-rr}
\a. The woman brought \ul{the sandwich from the kitchen} \textbf{tripped} on the carpet. [\textsc{reduced},~\textsc{ambig}uous]\label{ex:mv-rr-reduced-ambig}
\b. The woman who was brought \ul{the sandwich from the kitchen} \textbf{tripped} on the carpet. [\textsc{unreduced},~\textsc{ambig}]\label{ex:mv-rr-unreduced-ambig}
\c. The woman given \ul{the sandwich from the kitchen} \textbf{tripped} on the carpet. [\textsc{reduced},~\textsc{unambig}uous]\label{ex:mv-rr-reduced-unambig}
\d. The woman who was given \ul{the sandwich from the kitchen} \textbf{tripped} on the carpet. [\textsc{unreduced},~\textsc{unambig}] \label{ex:mv-rr-unreduced-unambig}

} 

In Example~\ref{ex:mv-rr-reduced-ambig}, the verb ``brought'' is initially analyzed as a main verb phrase, but upon reaching the verb ``tripped''---the disambiguator in this case---the reader must re-analyze it as an RC. The garden path should be eliminated in sentences such as ~\ref{ex:mv-rr-unreduced-ambig}, the \textsc{unreduced} condition, where the words ``who was'' clarify that the verb ``brought'' is part of an RC, rather than the main verb of the sentence. Therefore we quantify the garden path effect as the surprisal at the disambiguator for the \textsc{reduced} minus \textsc{unreduced} conditions.

There is another possible cue that the initial verb is the head of an RC: the morphological form of the verb. In examples such as ~\ref{ex:mv-rr-reduced-unambig}, the the verb ``given'' is unambiguously in its past-participle form, indicating that it cannot be the main verb of the sentence. If a language model is sensitive to morphological cues to syntactic structure, then it should either not show a garden path effect in this \textsc{unambig}uous condition, or it should show a reduced garden path effect.


We constructed 29 experimental items following the template of~\ref{ex:mv-rr}. Figure~\ref{fig:mv-rr-surprisal} shows the garden path effect sizes by model and verb-form ambiguity. All networks show the basic garden path effect ($p<.001$ in \googlernn, \gulordavarnn, and \rnng; $p<0.01$ in \tinylstm). However, the garden path effect in \tinylstm is much smaller than the other models: RC reduction causes an additional .3 bits of surprisal at the disambiguating verb, as compared to 2.8 bits in the \rnng, 1.9 in \googlernn, and 3.6 in \gulordavarnn (\tinylstm's garden path effect is significantly smaller than each other model at $p<0.001$). 

If the network is using the morphological form of the verb as a cue to syntactic structure, then it should show the garden path effect more strongly in the \textsc{ambig} condition than the \textsc{unambig} condition. The large language models and the \rnng do show this pattern: at the critical main-clause verb, surprisal is superadditively highest in the reduced ambiguous condition (the dotted blue line; a positive interaction between the reduced and ambiguous conditions is significant in the three models at $p<0.001$). However, \tinylstm
does not show evidence for superadditive surprisal for the ambiguous verbform and the reduced RC ($p=.45$).

The three large LSTMs and the \rnng replicate the key human-like garden-path disambiguation effect due to to ambiguity in verb form.
But strikingly, even when the participial verbform is unambiguous, there is still a significant garden path effect in all models ($p<0.01$ in all models except \tinylstm, where $p=.08$). Apparently, these networks treat an unambiguous passive-participial verb as only a \emph{noisy cue} to the presence of an RC. 

\section{General Discussion and Conclusion}

In all models studied, we found clear evidence of basic incremental state syntactic representation. However, models varied in how well they fully captured the effects of such state and the potentially subtle lexical cues indicating the beginnings and endings of such states: only the large LSTMs could sometimes reliably infer clause boundaries from verb argument structure (Section~\ref{sec:npz}) and morphological verb-form (Section~\ref{sec:mv-rr}), and only \gulordavarnn and the \rnng fully captured the proper behavior of subordinate clauses. The results are summarized in Table~\ref{tab:overall-results}. We suggest that representation of course-grained syntactic structure requires either syntactic supervision or large data, while exploiting fine-grained lexical cues to structure requires large data.

More generally, we believe that the psycholinguistic methodology employed in this paper provides a valuable lens on the internal representations of black-box systems, and can form the basis for more systematic tests of the linguistic competence of NLP systems. We make all experimental items, results, and analysis scripts available online at \url{github.com/langprocgroup/nn\_syntactic\_state}.


\bibliographystyle{acl_natbib}
\bibliography{everything}

\end{document}